\documentclass[10pt]{article}

\usepackage{arxiv}
\usepackage{definition}

\title{Robust Graph Structure Learning with the Alignment of Features and Adjacency Matrix}

\author{Shaogao Lv\footnotemark[1] 
\and
Gang Wen\footnotemark[1] 
\and
Shiyu Liu\footnotemark[2] 
\and
Linsen Wei\footnotemark[3] 
\and
Ming Li\footnotemark[4] 
}

\begin{document}
\date{} 
\maketitle

\renewcommand{\thefootnote}{\fnsymbol{footnote}}
\footnotetext[1]{Department of Statistics and Data Science, Nanjing Audit University, China.}
\footnotetext[2]{University of Electronic Science and Technology of China, China.}
\footnotetext[3]{School of Astronautics, Northwestern Polytechnical University, China.}
\footnotetext[4]{Key Laboratory of Intelligent Education Technology and Application of Zhejiang Province, Zhejiang Normal University, Jinhua, China; Key Laboratory of Scientific and Engineering Computing (Ministry of Education), Shanghai Jiao Tong University, China.}
\renewcommand{\thefootnote}{\arabic{footnote}}

\begin{abstract}
    To improve the robustness of graph neural networks (GNN), graph structure learning (GSL) has attracted great interest due to the pervasiveness of noise in graph data. Many approaches have been proposed for GSL to jointly learn a clean graph structure and corresponding representations. 
    To extend the previous work, this paper proposes a novel regularized GSL approach, particularly with an alignment of feature information and graph information, which is motivated mainly by our derived lower bound of node-level  Rademacher complexity for GNNs. 
    Additionally, our proposed approach incorporates sparse dimensional reduction to leverage low-dimensional node features that are relevant to the graph structure. To evaluate the effectiveness of our approach, we conduct experiments on real-world graphs. The results demonstrate that our proposed GSL method outperforms several competitive baselines, especially in scenarios where the graph structures are heavily affected by noise. Overall, our research highlights the importance of integrating feature and graph information alignment in GSL, as inspired by our derived theoretical result, and showcases the superiority of our approach in handling noisy graph structures through comprehensive experiments on real-world datasets.
\end{abstract}

\section{Introduction}  \label{sec:intr}
Graph neural networks (GNNs) have received increasing attention in recent years and have achieved remarkable performance across various tasks, including node classification \citep{welling2016semi,xu2018powerful}, recommendation systems \citep{wu2019session,yu2020tagnn}, and information retrieval \citep{yu2021graph}. In essence, GNNs employ a message-passing framework, wherein node embeddings are derived through the aggregation and transformation of neighboring embeddings.

The graph structure in GNNs distinguishes it from traditional neural network models. The success of vanilla GNNs over graph data relies heavily on one fundamental assumption, i.e., the original graph structure is reliable \citep{zhou2020graph}. However, inherent noise in graph data often exists in graph structure due to measurement errors or adversarial attacks \citep{jin2021adversarial,zugner2019adversarial}. A variety of (graph) neural networks are susceptible to noise \citep{szegedy2013intriguing,dai2018adversarial,wu2019adversarial}, especially inevitable noise in graph data that significantly diminishes the quality of representations produced by deep GNN models. Consequently, it is impossible to apply noisy GNNs to risk-critical practical problems, such as finance management and medical analysis. Therefore, it is of significant importance to develop robust GNN models that can effectively counter adversarial attacks and mitigate the impact of noise.


 Graph structure learning (GSL), which aims to simultaneously learn an optimized graph structure and corresponding representations, has achieved considerable success in recent years. This is achieved by modifying the graph structure, particularly through operations such as adding, deleting, or rewiring edges \citep{zhu2021deep}. In previous related literature, a fundamental concern in developing efficient GSL methods for GNN models is to generate a ``clean'' graph structure for learning representations and downstream tasks. However, one controversial and challenging issue for GSL lies in the criteria for a clean graph structure. 

It is frequently observed that real-world graphs often share specific properties. For instance, many real-world clean graphs tend to possess characteristics such as low-rankness and sparsity \citep{jin2020graph,zhou2013learning}. However, previous studies have overlooked the inherent relationships between node features and the underlying graph structure. Taking social networks as an example, individuals with similar hobbies are more likely to be friends. To effectively integrate node features with the raw graph structure, a core idea for GSL is to learn an encoding function that assigns edge weights based on pairwise distances between node features or presentations. Subsequently, the final graph structure for GNNs is refined by incorporating both the original graph structure and the learned one. Existing studies mainly consist of various metric learning approaches for GSL \citep{li2018adaptive,wu2018quest,halcrow2020grale,zhang2020gnnguard}.

In this paper, our core idea for GSL is primarily inspired by a novel theoretical finding concerning the Rademacher complexity of GNNs, a widely used notation that can describe the generalization capacity of a learning algorithm. Specifically, \cref{lower} in \cref{sec:method} demonstrates that the lower bound of Rademacher complexity of GNNs relies on the alignment between the feature and graph information. Furthermore, it is worth considering that practical problems often involve high-dimensional node features, while not all variables are necessarily associated with the graph structure. Taking this into account, the distance learning method for GSL proposed in this paper takes into consideration both the low dimensionality and sparsity of node features. In summary, we design robust graph neural networks by embedding the alignment between the feature and graph information, as well as leveraging the feature properties of sparsity and low dimension.

\paragraph{Contributions.} 
In particular, this paper concentrates on semi-supervised classification models that utilize graph convolutions such as graph convolutional networks (GCNs, \citealt{welling2016semi}), a class of popular graph models that bridge the gap between spectral and spatial domains. Our method is also applicable to both unsupervised and supervised graph models. For notational simplicity, we assume the node feature is noiseless. Overall, our contributions to GSL can be summarized as follows:

\begin{itemize}[leftmargin=*]
    \item We first establish a minimax lower bound of node-level Rademacher complexity of GCNs. This theoretical result provides valuable insights into the complexity of GCN models and serves as a foundation for our subsequent work. Building on this theoretical finding, we further propose a novel robust graph neural network for GSL. Our method takes into account the degree of alignment between the graph structure and node features, as well as the low dimensionality and sparsity of node features. By considering these factors, we aim to enhance the performance and robustness of GNN models in the context of GSL. 
    
    \item A suitable graph structure is a fundamental prerequisite for the success of GNNs. To the best of our knowledge, we are the first to consider the alignment between the graph structure and node features in the GSL literature. By addressing this crucial aspect, we contribute to filling a gap in the existing literature and advancing the understanding of GSL methods. 
    
    \item We conduct extensive experiments on three real-world datasets to evaluate the effectiveness of our proposed method. The results demonstrate the superiority of our method, further validating its effectiveness in GSL tasks against several competitive methods. 
\end{itemize}

\paragraph{Organization.} 
The rest of the paper is organized as follows. \cref{sec:related_work} reviews some of the related work. \cref{sec:method} presents several basic notations used on GNNs and introduces the theoretical results concerning the Rademacher complexity of GCNs. We further outline the methodology for GSL. Additionally, we present an alternative optimization approach for our proposed method in \cref{sec:algorithm}. \cref{sec:expr} presents the experiment results of our proposed method. Finally, we conclude the work in \cref{sec:conclusion}.



\paragraph{Notation.} 
The notation used in the paper is as follows:
For a vector $\ba$, $\|\ba\|_2$ refers to the standard norm in Euclidean space.
For a matrix $\bZ$, $\|\bZ\|_2$ and $\|\bZ\|_F$ denote the spectral norm and Frobenius norm of $\bZ$, respectively.
These notations allow for a clear and concise representation of vector and matrix norms, which are essential in the theoretical analysis and formulation of the proposed methods in the paper.

\section{Related Work} \label{sec:related_work}

\paragraph{Graph Neural Networks.}  
Common GNNs are divided into two categories: spectral GNNs and spatial GNNs (see \citep{wu2020comprehensive} for a related survey). In graph spectral theory,  spectral GNNs are a kind of GNN that designs graph signal filters in the spectral domain. For example, based on the graph Laplacian, \citet{bruna2013spectral} propose the graph convolution operation in the Fourier domain. \citet{defferrard2016convolutional} use Chebyshev polynomials as the convolution filter. \citet{welling2016semi} utilize the first-order approximation of Chebyshev that achieves a fast approximate convolution on graphs. \citet{wu2019simplifying} reduce the excess complexity of graph convolution into a single linear model, which achieves state-of-the-art performance in overall predictive accuracy.

On the other hand, using adjacency neighbors, spatial GNNs can intuitively define convolution operations on graphs. Specially, \citet{velivckovic2017graph} utilize an attention mechanism to aggregate representations of neighbors. \citet{hamilton2017inductive} first sample the neighbors and then aggregate the information to generalize the graph convolution. \citet{chen2018fastgcn} implement importance sampling on each convolutional layer, which improves learning efficiency.

\paragraph{Graph Structure Learning.}   
To alleviate the dependence of learning GNNs on the graph structure, recent efforts have been made for GSL (see \citep{zhu2021deep} for a related survey), which learn the graph structure and GNN parameters jointly. 
In particular, \citet{franceschi2019learning} model each edge in the adjacency matrix as a parameter, and learned these together with the GNN parameters in a two-level way. \citet{jiang2019semi} propose a similarity-based GSL with the help of node features. \citet{chen2019deep} learn the metrics to generate the graph structure between node features and GNN embeddings iteratively. \citet{jin2020graph} jointly learn a structural graph and a robust graph network model with graph properties. However, these aforementioned GSL methods do not fully consider the matching between node features and graph structure.


\section{Methodology} \label{sec:method}
\subsection{Preliminaries}
Consider an undirected network or graph $\mathcal{G}=(\mathcal{V},\mathcal{E})$, where $\mathcal{V}=\{\nu_1,\nu_2,...,\nu_n \}$ denotes the set of nodes with $|\mathcal{V}|=n$ and $\mathcal{E}$ represents the set of edges between nodes. The edge in the graph quantifies certain relationships among the data, such as correlation, similarities, or causal dependencies. The graph can be equivalently described by a (weight) adjacency matrix $\bA \in \mathbb{R}^{n \times n}$. In this paper, we focus on the standard binary setting where $\bA_{ij}=1$ indicates the existence of an edge between node $i$ and node $j$, otherwise $\bA_{ij}=0$.   

In addition to the edge information over the graph, the typical supervised setting assigns a pairwise feature $(\bx,y) \in \mathbb{R}^d\times \mathcal{C}$ at each node. The main task of such graph learning is to learn a function $f_\Theta: \mathcal{V}\rightarrow \mathcal{C}$, parameterized by $\Theta$, which fully exploits the underlying pattern based on useful information of the graph structure and the other regular samples. Consider an input matrix $\bX = (\bx_1, \bx_2,..., \bx_n)^T \in \mathbb{R}^{n\times d}$, where $\bx_i$ represents the attribute feature of node $i$.
In a semi-supervised setting,  only some labels of nodes can be observed, denoted by $\mathcal{V}_m=\{\nu_1,\nu_2,...,\nu_m\}$ with $m\ll n$, and the corresponding labels are given by $\by_L=\{y_1,y_2,...,y_m\}$. The objective function for GNNs can be formulated as:
\begin{align} \label{orgreg}
    \min_{\Theta}\mathcal{L}_{gnn}(\Theta,\bX,\by_{m}, \bA) ={\sum}_{\nu_i\in \mathcal{V}_m}\ell\big(f_\Theta(\bX,\bA)_i,y_i\big),
\end{align}
where $\ell:\mathbb{R}\times\mathbb{R}\rightarrow \mathbb{R}^+$ is a loss function, usually, the least square function is used for regression, while the cross entropy is used for classification. 
For clarity,  this paper focuses on GCNs originally introduced in \citep{welling2016semi}, although
our idea can be easily extended to other GNNs.
Specifically, a two-layer GCN with $\Theta = (\bW_1, \bW_2)$ implements $f_\Theta$ as
\begin{align} \label{functionform}
    f_{\Theta}(\bX,\bA)=\sigma_2\big(\bar{\bA}\sigma_1(\bar{\bA}\bX\bW_1)\bW_2\big),
\end{align}
where $\bar{\bA}=\bar{\bD}^{-1/2}(\bA+I)\bar{\bD}^{-1/2}$, $\bar{\bD}$ is the diagonal matrix of $\bA+I$
and $\sigma_1,\sigma_2$ are two activation functions such as ReLU and softmax. 

In this paper, we are primarily interested in the case where the available graph matrix $\bA$ is full of noise, due to measurement error or adversarial attack. It is known that existing standard GNNs are sensitive to noise, and even a small amount of noise in the graph can propagate to neighboring nodes, impacting the embeddings of many nodes. To improve the robustness of GNN models, one key idea is to produce a denoised graph structure that can be used for learning representations. 

As previously discussed, to deal with the GLS problem, most of the existing work imposed prior information for $\Theta$ as one  regularization of \cref{orgreg}, such as low rank assumption and sparse structure \citep{jin2020graph}. 


Within the framework of empirical risk minimization augmented with a regularization term, we introduce a novel regularization term that captures the alignment between the graph structure and node features. This idea is motivated by both an upper bound and an additional lower bound on the generalization performance of GNNs, which will be discussed in detail in the subsequent subsection.

\subsection{Motivation from Generalization Bounds}
A predictor with a generalization guarantee is closely related to the complexity of its hypothesis space.  We adopt (empirical) Rademacher complexity to measure the functional complexity, which can be used to directly obtain one generalization error. For a function set $\mathcal{F}$ defined over graph $\mathcal{G}$, the empirical Rademacher complexity is defined as
$$
\widehat{\mathcal{R}}(\mathcal{F}):=\mathbb{E}_{\epsilon}\Big[\frac{1}{m}\sup_{f\in \mathcal{F}}\Big|{\sum}_{j=1}^m\epsilon_j f(\bx_j)\Big|\bx_1,\bx_2,...,\bx_N\Big],
$$
where $\{\epsilon_i\}_{i=1}^m$ is an i.i.d. family (independent of $\bx_i$) of Rademacher variables. Note that the conditional expectation here is taken with respect to $\{\epsilon_i\}_{i=1}^m$ given that $\{\bx_i\}_{i=1}^n$ is fixed and is not limited to the supervised input data over $\mathcal{V}_m$.

Since the neighbor representation of graph shift operators is maintained, the $t$-th output of the first layer in \cref{functionform} can be written as a vector form
\begin{align}\label{fisrtequ}
    \sigma\Big({\sum}_{l=1}^d \bw_{l}^{(t)}{\sum}_{j=1}^n\bar{\bA}_{\nu j}x_{jl}\Big)=\sigma\Big({\sum}_{j\in N(\nu)}\bar{\bA}_{\nu j} \big\langle\bx_j, \bw^{(t)} \big\rangle \Big),
\end{align}
where $N(\nu)$ denotes the set of neighbors of $\nu$, and $\bW_1=(\bw^{(1)},\bw^{(2)},...\bw^{(k)})\in \mathbb{R}^{d\times k}$ is represented in a column-wise manner.  Note that we write $\sigma=\sigma_1=\sigma_2$ for notional simplicity. 

Thus, the class of functions defined over the  node set $\mathcal{V}_L$  with norm constraints  coincides with
\begin{align}\label{hypothesis}
    \mathcal{F}_{D,R}:=\Big\{f(\bx_i)
    &=\sigma\big(\sum_{t=1}^kw_2^{(t)}\sum_{\nu=1}^n\bar{\bA}_{iv}\times \sigma\big(\sum_{j\in N(\nu)}\bar{\bA}_{\nu j} \big\langle\bx_j, \bw^{(t)} \big\rangle \big) \big),
    \,i\in[L], 
    \; \|\bW_1\|_{F}\leq R,\,\|\bW_2\|_2\leq D \Big\},
\end{align}
where the Frobenius norm  of a matrix  is given as $\|\bW\|_{F}^2:=\sum_{ij}W_{ij}^2=\sum_{t=1}^k\|\bw^{(t)}\|_2^2$.  Bounding the population Rademacher complexity over  $\mathcal{F}_{D,R}$ is quite challenging, mainly due to the fact that each output $h_i^{(2)}$ depends on all the input features that are connected to node $\nu_i$, as shown in  \cref{functionform,hypothesis}. Although some upper bounds of Rademacher complexity for specific GNNs have been provided in \citep{esser2021learning,garg2020session}, it may be argued that these upper bounds are suboptimal in some senses, therefore, there is a lack of explainability and persuasiveness. 
As a necessary supplement, we now provide a minimax lower bound of Rademacher complexity over a two-layer GCN, to reveal some essential factors together with their tight upper bounds stated in \cref{sec:background}. 

Without loss of generality, we assume that the number of neighbors is equal for all nodes, denoted by $q$. 
Let $\widetilde{\mathbf{X}}_v=\left(\tilde{\mathbf{x}}_1^T, \ldots, \tilde{\mathbf{x}}_q^T\right)^T \in \mathbb{R}^{q \times d}$ be the feature matrix of the nodes in $\mathcal{G}_v$, where all $\tilde{\mathbf{x}}_i$ 's are denoted to be reordered input data according to the neighbors of node $\nu$. We are state our main results.
\begin{thm}[Lower Bound of Rademacher Complexity] \label{lower}
	Let $\mathcal{F}_{D,R}$ be a class of GCNs with one hidden layer, where the parameter matrix and the parameter vector satisfy 
	$\|\bW_1\|_F\leq R$  and $\|\bW_2\|_2\leq D$ respectively. Then there exists a choice of $l$-Lipschitz activation function, data points $\{\bx_{i}\}_{i=1}^n$ and a family of given graph convolutional filters, such that
	$$
	\widehat{\mathcal{R}}(\mathcal{F}_{D,R})\geq
	\frac{l^2BDR}{\sqrt{m}}\min_{k\in [q]}\Big\{ \big\|\widetilde{\bX}_q\bar{\bA}_{\cdot k} \big\|_2\sum_{t=1}^q\bar{\bA}_{kt}\Big\},
	$$
 where the constant $B:=\max_{i\in [N]} \|\bx_i\|_2$. 
\end{thm}
\cref{lower} indicates that the lower bound of $\widehat{\mathcal{R}}(\mathcal{F}_{D,R})$ depends on the number of labels, the degree distribution of the graph, and the choice of the graph convolution filter. It is interesting to observe that, the above bound is independent of the graph size ($n$) in general.
It is also worth noting that, for the two-layer neural network with width $k$, our
lower bound only has an explicit dependence on the Frobenius norm of the parameter matrix,
while is independent of the network width. Importantly, the matching between the graph matrix and the features plays a crucial role in determining the best-case generalization performance of GCN.

\subsection{Method Formulation}
In this subsection, we use the metric learning approach \citep{li2018adaptive} to update the graph structure based on the input features. This involves deriving edge weights through learning a metric function that measures the pairwise similarity of representations.
We define a nonnegative function $\phi: \mathcal{X}\times \mathcal{X} \rightarrow \mathbb{R}^+$ between data $\bx_i$ and $\bx_j$ by
\begin{subequations}
\begin{align}
    \phi(\bx_i,\bx_j)&=\sqrt{(\ba\circ(\bx_i-\bx_j))^T\bM^T \bM \ba\circ(\bx_i-\bx_j)}, 
    \\
    \widetilde{\bA}_{ij}&=\exp\Big(-\frac{\phi(\bx_i,\bx_j)^2}{2\tau^2}\Big), \label{phixixj}
\end{align}
\end{subequations}
where the symbol $\circ$ denotes the element-wise multiplicative, namely, $\ba\circ \bx:=(\ba_1\bx^{(1)},...,\ba_d\bx^{(d)})$. The vector $\ba$ is a trainable parameter with a sparsity constraint, which allows us to select only a few relevant features for the graph structure. The matrix $\bM \in \mathbb{R}^{p \times d}$ is also trainable and used for projecting node embeddings into a latent space, where $p \leq d$ for dimension reduction. 

We introduce feature selection to the GSL process because not all features are necessarily related to the graph structure. In practical problems, such as chemical and molecular graphs, there can be strong heterophily, where certain features of connected nodes exhibit significant variation \citep{zhu2020beyond}. Our sparsity-based feature selection approach is different from previous sparse GSL methods \citep{jin2020graph}, where sparsity primarily refers to the adjacency matrix.

The learned feature-based matrix $\widetilde{\bA}$ is then combined with the original structure $\bA$ to form a new adjacency matrix in an interpolation manner:  
\begin{align}\label{newA}
    \widehat{\bA}=(1-\alpha)\bA+\alpha \widetilde{\bA},
\end{align}
where $\alpha\in[0,1]$ is a tuning hyperparameter that mediates the influence of the learned structure.
 
Based on the basic notation described above,  we introduce a metric-based regularizer by 
\begin{subequations}
\begin{align}
    \mathcal{L}_{\text{ss}}(\bM,\ba):&=\frac{1}{2}{\sum}_{i,j=1}^{n}\|\bx_i-\bx_j\|_2^2\widetilde{\bA}_{ij}+\lambda_1\|\ba\|_1 \\ 
    &=\operatorname{tr}(\bX^T(\widetilde{\bD}-\widetilde{\bA})\bX)+\lambda_1\|\ba\|_1, \label{ssreg}
\end{align}
\end{subequations}
where $\widetilde{\bD}$ is the diagonal matrix of $\widetilde{\bA}$. The first term in \cref{ssreg} is to enforce certain smoothness for graph structure \citep{jin2020graph}, while the second term is used to generate  sparse parameters associated with the input features.

In addition to the graph update based on some of the input features, it is worth mentioning that the alignment of features and graph plays a significantly positive role in the generation performance of general GNNs, as shown in our result in \cref{lower}, as well as an upper bound of transductive Rademacher complexity on GNNs \citep{esser2021learning}. In the minimax sense, as $\|\bX\bA\|_2$ becomes smaller, the corresponding minimax generalization error of various GNNs will be sharper.
Inspired by such a theoretical finding, we introduce another regularizer with respect to  the graph constraint:
\begin{align}\label{alignreg}
\mathcal{L}_{\text{align}}(\bM,\ba)=\|\bX\widehat{\bA}\|_2.
\end{align}
Therefore, our final objectives can be formulated by combining \cref{orgreg,ssreg,alignreg}
\begin{align}\label{combining}
    \min_{\Theta,\bM,\ba}\mathcal{L}_{\text{gnn}}(\Theta,\bX,\by_{L}, \widehat{\bA})
    &+\gamma_1\mathcal{L}_{\text{ss}}(\bM,\ba) 
    +\gamma_2\mathcal{L}_{\text{align}}(\bM,\ba). 
\end{align}

Given that  dense graphs not only lead to a heavy computational burden, they also might contain noise. Hence, it is common to prune edges according to edge weights as a post-processing operation, which results in either a $k$NN graph (i.e. each node has up to $k$ neighbors) or a $\epsilon$NN graph (i.e. edges whose weights are less than $\epsilon$ will be discarded). 


\subsection{Numerical Algorithm} \label{sec:algorithm}

Since it is difficult to optimize \cref{combining} directly,  we consider an alternate optimization algorithm associated with GNN parameters $\Theta$ and parameters $\bM, \ba$, respectively. Our proposed framework consists of the following two steps.

\paragraph{Step 1: update $(\bM,\ba)$.}
Note that the GNN is irrelevant to the parameters $(\bM,\ba)$, and thus we can update them without calculating GNN. Only the  empirical loss  involves the graph structure. Fix $\widehat{\bA}$, the objective function in \cref{combining} to update $(\bM,\ba)$ is represented as:
\begin{align}\label{combining1}
\begin{split}
    {\min}_{\bM,\ba} \; \mathcal{L}_{\text{gnn}}(\Theta,\bX,\by_{L}, \widehat{\bA}) +\gamma_1\mathcal{L}_{\text{ss}}(\bM,\ba)+\gamma_2\mathcal{L}_{\text{align}}(\bM,\ba).
\end{split}
\end{align}

\paragraph{Step 2: update $\Theta$.}
When we calculate the metric learning of node features and update the adjacency matrix of $\bA$ by $\hat{\bA}$, we can reduce \cref{combining} to the empirical loss:
\begin{align}\label{combining2}
    {\min}_{\Theta} \; \mathcal{L}_{\text{gnn}}(\Theta,\bX,\by_{L}, \widehat{\bA}).
\end{align}
      
      
Before presenting our algorithm, we introduce the following basic notations. The identity matrix is denoted by $\mathbf{E}$, $\mathbf I$ is a vector whose elements are 1.
 A function $f$ is a metric learning with the Gaussian kernel.
The $prox$ algorithm \citep{beck2009fast} is a forward-backward method designed to approximate a gradient at non-derivable points.
It can be presented as follows:
\begin{align}\label{prox}
    prox_{\eta \|\cdot\|_{1}} (\ba) = \operatorname{sgn}(\ba) \odot \max{\big(|\ba| - \lambda_1, 0\big)},
\end{align}
where $\eta>0$ is the learning rate, $\operatorname{sgn}$ is the sign function, $\lambda_1$ is a weight of $\ell_1$-norm, and $\odot$ is the Hadamard product. 

\begin{algorithm}
\caption{\textbf{R}obust \textbf{G}raph \textbf{S}tructure \textbf{L}earning with the \textbf{A}lignment of Features and Adjacency Matrix (\RGSLA)} 
\label{alg1}
\begin{algorithmic}[1]
\REQUIRE 
Adjacency matrix $\bA$, attribute matrix $\bX$, labels $\by_L$, 
tuning hyperparameters $\gamma_1, \gamma_2, \lambda_1, \alpha$, 
\\ \qquad 
window width $\tau$.
\ENSURE Learned adjacency $\hat{\bA}$, GNN parameter $\bTheta$.
\STATE \textbf{Initialize} $\widetilde{\bA} \leftarrow \bA,\; \bM \leftarrow \mathbf E,\; \ba\leftarrow \mathbf I$.
%
\FOR{$t=0$ to $T-1$}
\FOR{$i=0$ to $I$}
    \STATE $\widetilde{\mathbf{D}}^{(t)}, \widetilde{\bA}^{(t)} \leftarrow f(\bM^{(t)},\ba^{(t)},\bX,\tau)$.
    \STATE $\widehat{\bA}^{(t)}\leftarrow (1-\alpha)\bA + \alpha \widetilde{\bA}^{(t)}$.
    \FOR{$j=0$ to $J$}
        \STATE Update GCN parameters by
        $$
        \Theta^{(t+1)} \leftarrow \Theta^{(t)} - \eta_t \triangledown_{\Theta} \mathcal{L}_{\text{gnn}}(\Theta^{(t)},\bX,\by_{L}, \widehat{\bA}^{(t)}).
        $$
    \ENDFOR
    \STATE Update projecting matrix by
    \begin{align*}
        \bM^{(t+1)} \leftarrow &\bM^{(t)} - \eta_t^\prime \triangledown_{\bM} \big(\mathcal{L}_{\text{gnn}}(\Theta^{(t+1)},\bX,\by_{L}, \widehat{\bA}^{(t)})
        + \gamma_1 \operatorname{tr}\big(\bX^T(\widetilde{\bD}^{(t)}-\widetilde{\bA})^{(t)}\bX\big) 
        + \gamma_2 \mathcal{L}_{\text{align}} \big).
    \end{align*}
    \STATE Update sparse parameters by 
    $$
    \ba^{(t+1)}\leftarrow prox_{\eta \|\cdot\|_{1}} (\ba^{(t)}).
    $$
\ENDFOR
\ENDFOR
\end{algorithmic}
\end{algorithm}

\begin{figure*}[ht]
    \centering
    \subfigure[Cora]{
	\includegraphics[width=0.3\linewidth]{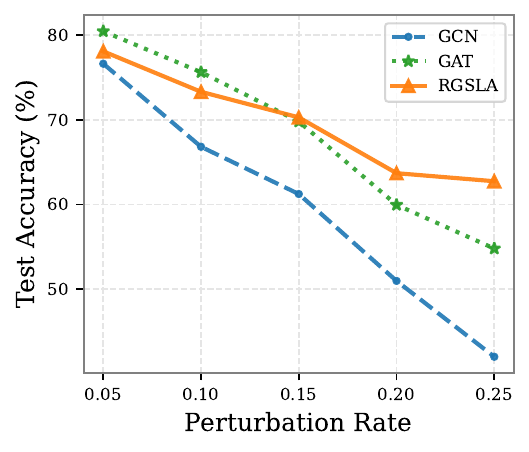}
	\label{fig:benchmark_cora}
    }
    \subfigure[Citeseer]{
        \includegraphics[width=0.3\linewidth]{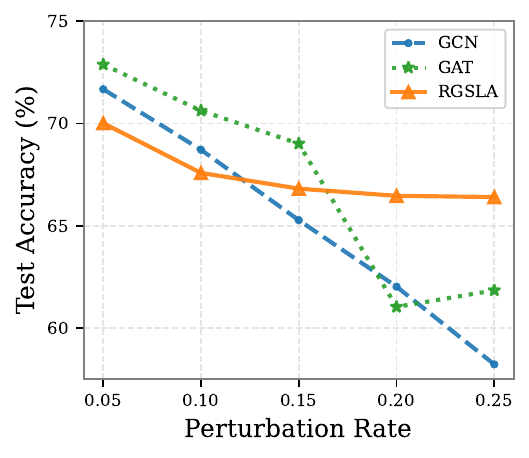}
        \label{fig:benchmark_citeseer}
    }
    \subfigure[Polblogs]{
        \includegraphics[width=0.3\linewidth]{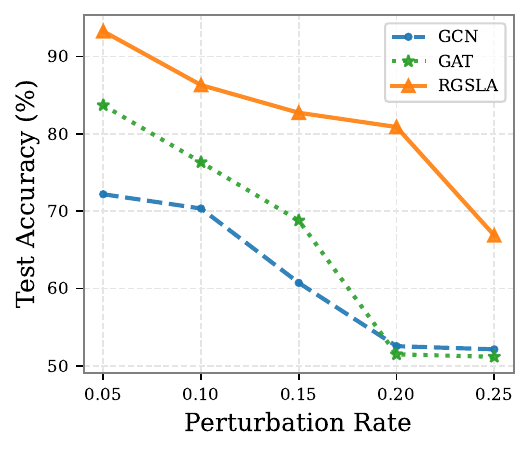}
        \label{fig:benchmark_polblogs}
    }
    \caption{Accuracy of different models under Metattack.}
    \label{fig:benchmark}
\end{figure*}


In \cref{alg1}, we start by initializing $\bM \leftarrow \mathbf E$ and $\ba \leftarrow \mathbf I$. Following that, we calculate the new adjacency matrix $\widehat{\bA}$ using \cref{newA} (Lines 4-5). Subsequently, the function parameter $\Theta$ of GNN is updated by calculating its gradient with respect to $(\bM, \ba)$ and $\widehat{\bA}$ through solving \cref{combining2} (Line 7). After that, both $\bM$ and $\ba$ are updated using gradient descent and the proximal operator, respectively, as described by \cref{combining1,prox} (Lines 8-9).


\section{Numerical Experiments} \label{sec:expr}

In this section, we use \emph{metattack} \citep{zugner2019adversarial}, which is an attack model to deteriorate the performance of the graph model, to poison the graph. Note that the goal of metattack is to launch a global attack so that it can reduce the overall classification accuracy of a model.

According to \citep{jin2020graph}, metattack connects nodes with a significant feature difference. So we can use metattack to poison the graph, and then we can use our model to learn the original graph structure based on the raw graph structure and node features.


\subsection{Experiment settings}
We employ the following real-world datasets to evaluate our proposed model. The statistics of these datasets are shown in \cref{tab:statistics}. 
\begin{itemize} [leftmargin=*]
    \item Cora \citep{mccallum2000automating} and Citeseer \citep{sen2008collective} are citation network datasets in which Cora has a total of 2708 nodes, and each node contains 1433 features. The value of each feature is 0/1, to indicate whether it contains feature words. Cora has a total of 5429 edges and 7 classifications.
    \item Citeseer \citep{sen2008collective} has a total of 3312 nodes. Each node contains 3703 features. Similar to Cora, the value of each feature is also 0/1. Citeseer has a total of 4732 edges and 6 classifications.
    \item Polblogs \citep{adamic2005political} is a political dataset with 1490 blog pages. Each blog page can be regarded as a node of the graph, and the connections between nodes are the connections between blogs. There are a total of 19,090 edges. The node labels of the dataset are conservative or liberal.
\end{itemize}

\begin{table}[th]
    \centering
    \caption{Summary of Dataset Statistics.}
    \label{tab:statistics}
    \begin{tabular}{c|rrrr}
    \toprule
    Dataset     &Nodes  &Features   &Labels &Edges \\
    \midrule
    Cora        &2708   &1143   &7  &5429 \\
    Citeseer    &3327   &3703   &6  &4732 \\
    Polblogs    &1490   &1222   &2  &19090 \\
    \bottomrule
    \end{tabular}
\end{table}

To assess the proposed  method, we conduct a comparative analysis against state-of-the-art GNN models using the DeepRobust adversarial attack repository. Among the various models of Graph Convolutional Networks (GCN), we specifically focus on the most prominent and representative one \citep{welling2016semi}. Graph Attention Network (GAT, \citealt{velivckovic2017graph}) is a network architecture that consists of attention layers, allowing it to learn different weights for different nodes in the neighborhood. GAT is commonly used as a baseline method for defending against adversarial attacks.We employ the metattack method, which is a representative non-targeted attack. This type of attack aims to diminish the overall performance of the GNN model on the entire graph, rather than targeting specific nodes or classes.
The perturbation method of the adjacency matrix of the graph is mettack \citep{zugner2019adversarial}, and the interference rate is 5\% to 25\%, in 5\% increments. All the experiments are conducted 10 times with different initializations and random seeds.

\subsection{Performance Comparison}
In \cref{fig:benchmark_cora}, we observe that our proposed method, RGSLA, outperforms both \GCN and \GAT, particularly when the interference rate is relatively high. Notably, when the interference rate exceeds 0.15, our model demonstrates superior robustness compared to other models.
For the Citeseer dataset in \cref{fig:benchmark_citeseer}, our model initially performs worse than \GCN and \GAT. However, as the interference rate of the adjacency matrix increases, both \GCN and \GAT aggregate the information of mislabeled nodes into the prediction nodes. This leads to a sharp drop in algorithm performance. In contrast, our proposed model can learn the connection relationships of the nodes through their features. Therefore, when the interference rate increases, our model can correct the adjacency matrix of the graph. Even when the interference rate is significant, our model's performance remains close to the uninterrupted state.
Moving on to the Polblogs dataset illustrated in \cref{fig:benchmark_polblogs}, our proposed model consistently outperforms both \GCN and \GAT due to its superior ability to correct the graph structure.


\begin{figure}[hb]
    \centering
    \includegraphics[width=0.4\linewidth]{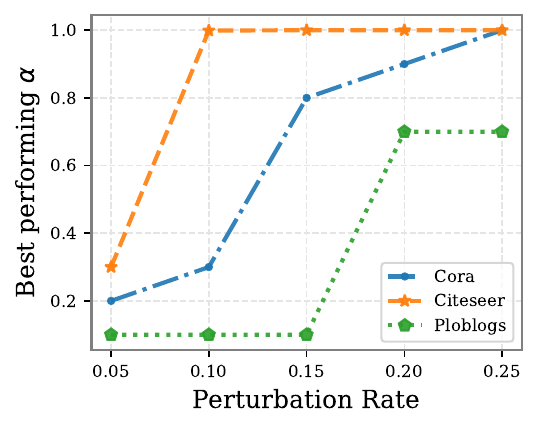}
    \caption{The optimal $\alpha$ varies with the perturbation rate.}
    \label{fig:alpha-ptb}
\end{figure}

\subsection{Ablation Study}
To gain a deeper understanding of the contributions of various components in defending against adversarial attacks, we conducted an ablation study. In our model, the parameter $\alpha$ plays a crucial role in mediating the influence of the learned structure on the model, transitioning from the input features of the original adjacency matrix. A higher value of $\alpha$ indicates a stronger inclination towards learning the structure directly from the input features to construct a new adjacency matrix.
To investigate the impact of the parameter $\alpha$ on the model, we compared the corresponding optimal values of $\alpha$ under different levels of noise disturbance, as depicted in \cref{fig:alpha-ptb}. The results show that for different datasets, when the data is heavily disturbed, a larger value of $\alpha$ leads to improved accuracy of the model. This observation confirms that on datasets with significant levels of contamination, the structure learned by the model directly from its features tends to yield better performance.


\subsection{Parameter Sensitivity}
We considered four crucial parameters in our study: $\alpha$, $\tau$, $\gamma_1$, and $\gamma_2$. Each of these parameters plays a significant role in controlling specific aspects of our model, including the influence of the learned graph, the distance scale, the smoothness of features, and the sparsity and alignment of the model.

To evaluate the impact of each component, we conducted experiments where we systematically varied the value of one parameter while keeping the other parameters fixed at zero. By doing so, we were able to observe how changes in a particular parameter affected the overall performance of the model. Analyzing the performance variations resulting from these experiments allowed us to gain a better understanding of the individual effects of each component. This knowledge helped us assess the importance of different parameters and their contributions to the overall effectiveness of our model.

\begin{figure}[th]
    \centering
    \includegraphics[width=0.4\linewidth]{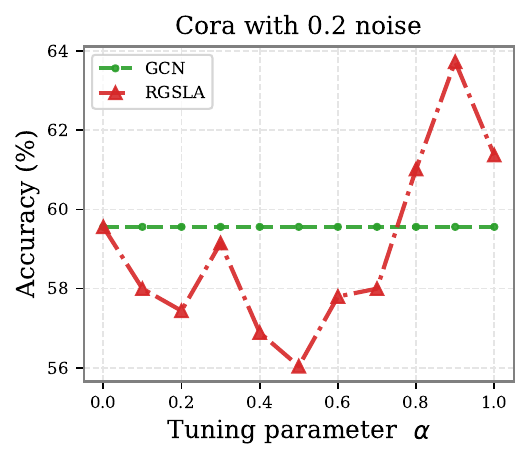}
    \caption{The sensitivity of tuning parameter $\alpha$.}
    \label{fig:sensty_alpha}
\end{figure}

\paragraph{Weight Parameter $\alpha$.}
Parameter $\alpha$ in this experiment represents the relative weight assigned to the original adjacency matrix and the similarity matrix learned from the graph structure. A larger value of $\alpha$ indicates a higher degree of reliance on the node features of the graph to compensate for the noise present in the graph structure. Specifically, when $\alpha$ is set to 0, the model is solely updated based on the adjacency matrix of the graph. Conversely, when $\alpha$ is set to 1, the model exclusively utilizes the information from the similarity matrix derived from the nodes of the graph. A larger value of $\alpha$ indicates a greater utilization of the information captured in the learned similarity matrix.

\begin{figure}[bh]
    \centering
    \includegraphics[width=0.4\linewidth]{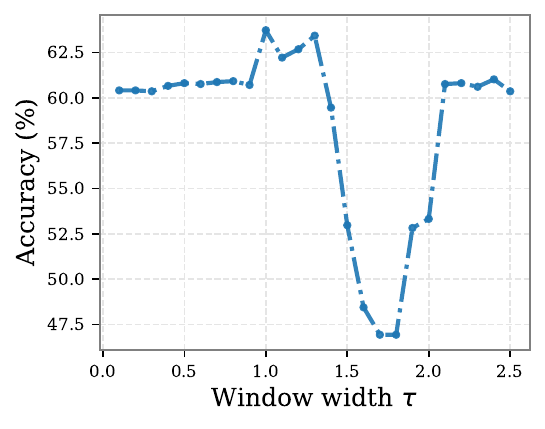}
    \caption{The sensitivity of window width $\tau$.}
    \label{fig:sensty_tau}
\end{figure}

\cref{fig:sensty_alpha} demonstrates the effect of different values of $\alpha$. When $\alpha=0$, the model is directly updated from the adjacency matrix, resulting in performance consistent with \GCN. As $\alpha$ gradually increases, the model begins to incorporate noise information from the nodes, which initially leads to poorer performance at lower $\alpha$ values. Due to the relatively high noise level in the data, smaller $\alpha$ values are insufficient for noise reduction. However, as observed in \cref{fig:sensty_alpha}, the model's performance gradually improves as $\alpha$ increases. Notably, when $\alpha=0.9$, the model significantly outperforms other models. This indicates that, when confronted with high levels of noise, the model effectively leverages information directly from the node features.

\begin{figure}[th]
    \centering
    \includegraphics[width=0.45\linewidth]{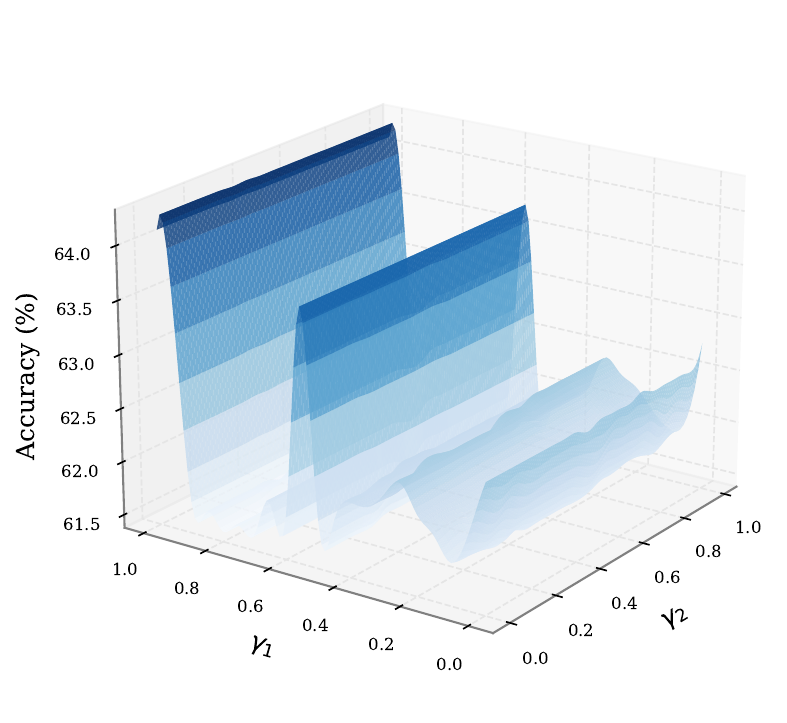}
    \caption{The sensitivity of regularized parameters $\gamma_1$ and $\gamma_2$.}
    \label{fig:sensty_gamma12}
\end{figure}

\paragraph{Window Width $\tau$.}
The size of the node distance is controlled by $\tau$. Using a Gaussian kernel function, a larger value of $\tau$ increases the likelihood of connecting nodes with greater distances. At the extreme, when $\tau$ is very large, all node connections have a weight of 1. Conversely, when $\tau$ is very small, the calculated weights are 0. In essence, the value of $\tau$ determines the extent of node connectivity, and different datasets may require different values of $\tau$. To illustrate the impact of variable $\tau$, we use the Cora dataset as an example, although similar observations hold for other datasets.

\cref{fig:sensty_tau} demonstrates that when $\tau$ is either too small or too large, the model achieves an accuracy of approximately 0.6. In these extreme cases, the adjacency matrix is either all 0s or all 1s, disregarding the original adjacency matrix in the calculations, resulting in similar outcomes. An inappropriate $\tau$ value can lead to erroneous node connections, where nodes with significant dissimilarities become connected, consequently interfering with the model's performance. Conversely, an appropriate choice of $\tau$ can improve the model's performance. Therefore, caution must be exercised when selecting the hyperparameter $\tau$.

\paragraph{The Regularized Parameters $(\gamma_1,\gamma_2)$.}
The two parameters, $\gamma_1$ and $\gamma_2$, control the regularized loss function. Specifically, $\gamma_1$ regulates the learned similarity between nodes. Nodes that are closer together are more likely to belong to the same class, while nodes that are farther apart are less likely to be connected. On the other hand, $\gamma_2$ governs the alignment between attribute features and the graph structure. When $\|\bX\bA\|_2$ is small, the generalization error of A given a GNN becomes sharper.

From \cref{fig:sensty_gamma12}, it can be observed that both $\gamma_1$ and $\gamma_2$ significantly impact the experiment accuracy. Considering the influence of the distance factor, $\gamma_1$ has a more pronounced effect on accuracy, while $\gamma_2$ also plays a role, albeit to a lesser extent compared to $\gamma_1$.



\paragraph{The Item Parameters $(\theta_1, \theta_2, \theta_3)$.}
Recall our loss function,
\begin{align*}
    & \mathcal{L}_{\text{gnn}} + \gamma_1\mathcal{L}_{\text{ss}}(\bM,\ba)+\gamma_2\mathcal{L}_{\text{align}}(\bM,\ba) 
    \\
    =& \mathcal{L}_{\text{gnn}} + \gamma_1 \hbox{tr}(\bX^T(\widetilde{\bD}-\widetilde{\bA})\bX) 
              + \gamma_1 \lambda\|\ba\|_1
              + \lambda_2 \mathcal{L}_{align}(\bM,\ba).
\end{align*}
The parameters $\gamma_1, \gamma_2, \lambda$ can be combined as new parameters $\theta_1, \theta_2, \theta_3$. Transformed by the new parameter, we can find the contribution of the three items of our model. Specifically,
\begin{align*}
\mathcal{L}_{\text{gnn}} + \theta_1 \hbox{tr}(\bX^T(\widetilde{\bD}-\widetilde{\bA})\bX) + \theta_2 \|\ba\|_1 + \theta_3 \|\bX\widehat{\bA}\|_2.
\end{align*}

\begin{figure*}[ht]
	\centering
	\subfigure[$\theta_1$ and $\theta_2$]{
	    \centering
		\includegraphics[width=0.3\linewidth]{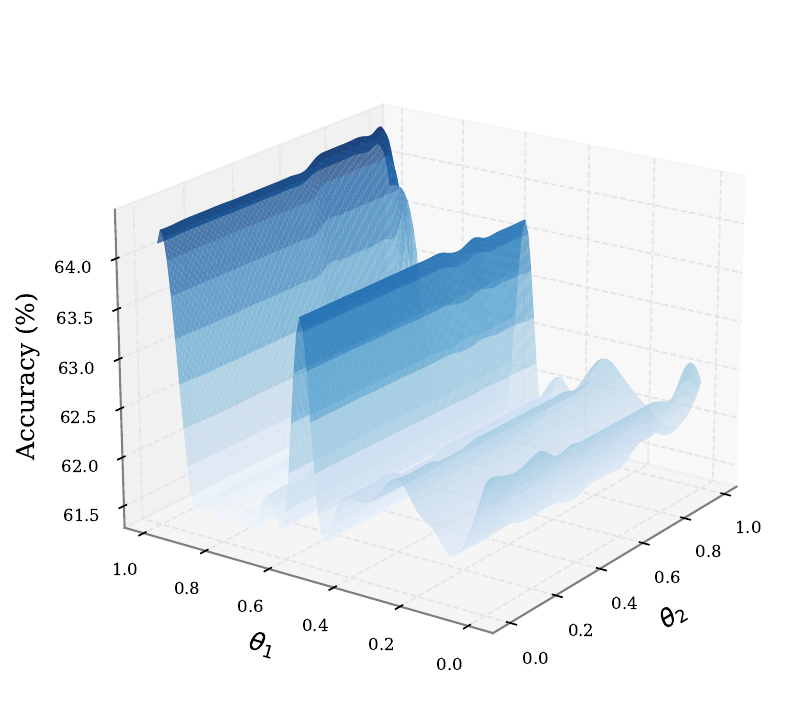}
            \label{fig:theta12}
	}
	\subfigure[$\theta_2$ and $\theta_3$]{
	    \centering
		\includegraphics[width=0.3\linewidth]{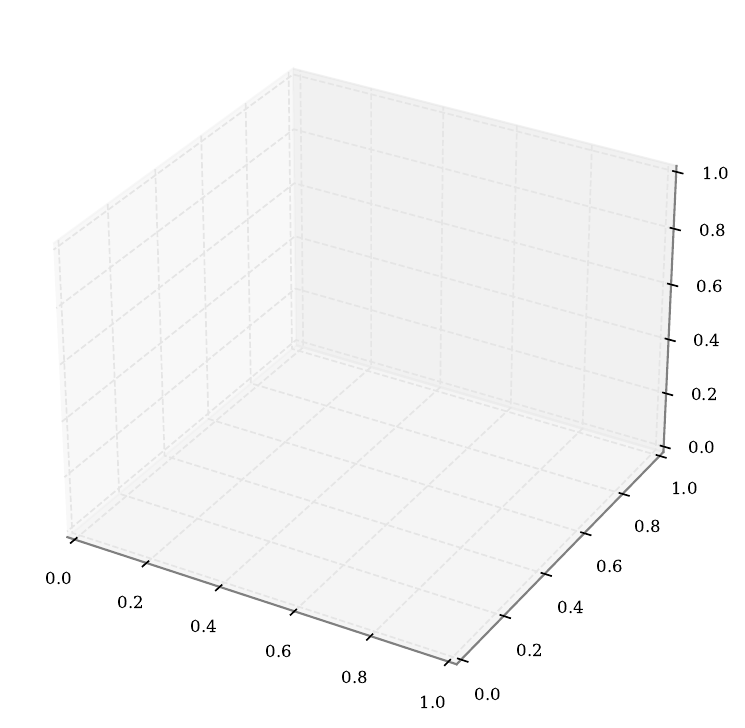}
            \label{fig:theta23}
	}
	\subfigure[$\theta_1$ and $\theta_3$]{
	    \centering
		\includegraphics[width=0.3\linewidth]{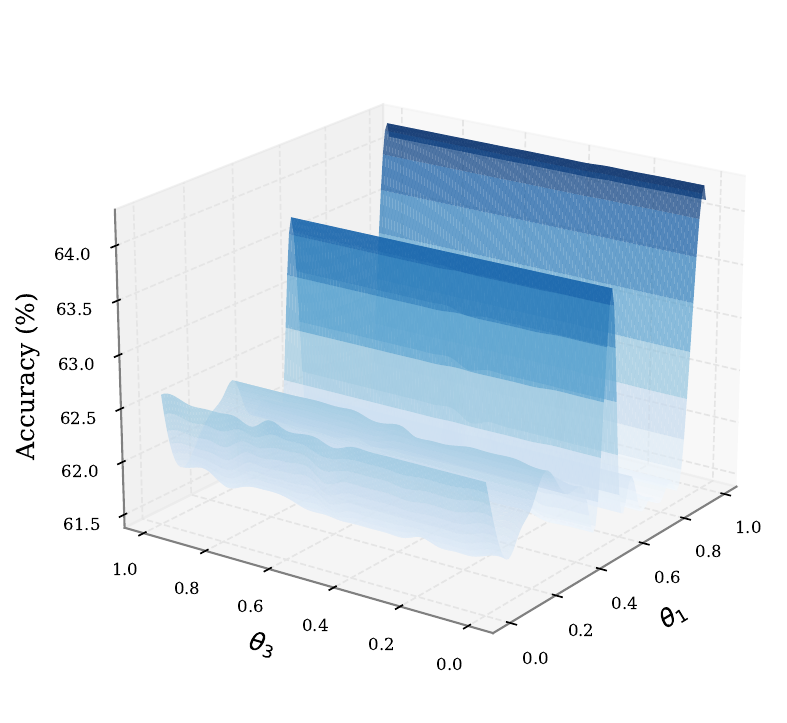}
            \label{fig:theta13}
	}
	\caption{The sensitivity of item parameters $\theta$.}
	\label{fig:theta123}
\end{figure*}

We define $\theta_1 = \gamma_1$, $\theta_2 = \gamma_1 \lambda$, and $\theta_3 = \gamma_2$. Subsequently, we set each of these parameters to zero individually and observe the influence of the remaining two parameters. As depicted in \cref{fig:theta123}, we can observe the impact of the parameter $\theta$ on our model. \cref{fig:theta12,fig:theta13} demonstrate that parameter $\theta_1$ has a significant effect on accuracy, as it controls the smoothness of the graph structure. On the other hand, $\theta_2$ and $\theta_3$ govern the sparsity of the input features and the relationship between the input feature and the adjacency matrix. Therefore, we can conclude that the smoothness of the graph structure is crucial for the GNN model, while the other two components also contribute to improved accuracy.



\subsection{Visualization of Noise Improvement}

This section investigates the robustness of the proposed model to noise in the adjacency matrix of graph-structured data. Such noise can arise from adversarial attacks or inherent imperfections in the graph itself. It is crucial to assess whether our proposed Graph Structure Learning (GSL) model remains effective in the presence of noise, and whether the estimated adjacency matrix accurately reflects the node connection information.
\begin{align}
    \bA \stackrel{\text{noise}}{\longrightarrow} \bA_{\text{noise}} \stackrel{\text{noise reduction}}{\longrightarrow} \widehat{\bA}
\end{align}



\begin{figure}[!h]
    \centering
    \includegraphics[width=0.7\linewidth]{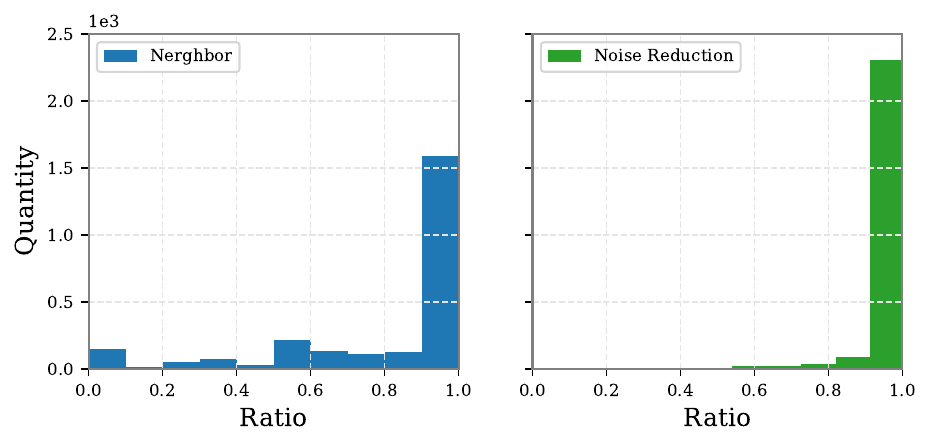}
    \caption{The improvement on noise reduction}
    \label{fig:6}
\end{figure}

Our method aims to capture more aggregated information from nodes with similar features. However, it is important to verify whether nodes with similar features, as determined by our model, also tend to share the same labels. \cref{fig:6} provides insights into this aspect.

For graph data, we calculate the ratio of nodes with the same label in the neighborhood connected to each node. A higher ratio indicates a larger presence of nodes with the same label in the neighborhood. Specifically, for a given node $j$, this ratio is defined as follows:
\begin{align}
    r_j = \frac{\text{the number of neighborhoods with the same label }}{\text{the number of all neighbors}}
\end{align}

By calculating the ratio for each node in the adjacency matrix and plotting the corresponding frequency distribution histogram, we can observe the distribution of same-label ratios in the neighborhood. In this study, we utilize the Cora dataset with a pollution rate set at 0.25. The left side of \cref{fig:6} shows the ratio distribution histogram for the original adjacency matrix of the graph data, while the right side displays the ratio distribution histogram for the adjacency matrix improved by our model.

As depicted in \cref{fig:6}, the ratio of nodes with the same label in the neighborhood is low in the original adjacency matrix of the graph data. However, after applying our model, the ratio of nodes with the same label increases. This allows the GNN to aggregate more data with the same label during the aggregation process, thereby reducing errors and enhancing the overall performance of the model.




\section{Conclusion} \label{sec:conclusion}
In this paper, we study GSL for GNNs and propose a novel robust GSL approach that simultaneously learns the graph structure and the GNN parameters. Remarkably, our proposed approach in the context of GSL is the first to consider the alignment of node features and graph structure. Such an idea is motivated by our derived lower bound of  empirical Rademacher complexity on GCN. Our experiments indicate that our approach mostly outperforms several competitive baselines 
and improves overall robustness under various amounts of noise pollution. In future research, several avenues can be explored to further advance the field of GSL and improve upon our proposed robust GSL approach: i) Enhancing model interpretability. It is meaningful to investigate advanced methods to enhance the interpretability of the learned graph structure and GNN parameters; ii) Handling dynamic graphs. It is strongly anticipated that we will further extend our proposed robust GSL approach to handle dynamic graphs where the graph structure evolves over time; iii) Transfer learning and generalization. The exploration of techniques to leverage pre-trained graph structure models or transfer knowledge from related tasks has good potential to enhance the performance and generalization capabilities of the robust GSL approach.

\section*{Acknowledgment}
Shaogao’s work is partially supported by the National Natural Science Foundation of China (No.11871277), and Young and Middle-aged Academic Leaders in Jiangsu QingLan Project (2022). Ming Li acknowledged the support from the National Natural Science Foundation of China (No. 62172370, No. U21A20473), the support from Zhejiang Provincial Natural Science Foundation (No. LY22F020004), and the support from the Fundamental Research Funds for the Central Universities.

\clearpage
\begin{small}
\bibliographystyle{plainnat}
\bibliography{main}
\end{small}

\clearpage

\begin{appendices}
\crefalias{section}{appendix}
\crefalias{subsection}{appendix}

The appendices are structured as follows. In \cref{sec:derivation}, we include additional derivation of the metric function between pairwise representations. In \cref{sec:background}, we provide the necessary backgrounds for our theoretical results. The proof of \cref{lower} is sketched in \cref{sec:proof}. Before delving into the appendices, we will first review the necessary notations used in the paper (refer to \cref{tab:notions}), which enable a clear and concise representation of the concepts and results discussed.

\begin{spacing}{1.05}
\listofappendices
\end{spacing}

\begin{table}[th]
    \centering
    \caption{Notations and their meanings in this paper.}
    \label{tab:notions}
    \begin{tabular}{l|l}
    \toprule
    Notations     & Meanings \\
    \midrule
    $\bX$       & the input matrix \\
    $\bx_i$     & the attribute feature of node $i$ \\
    $y_i$       & the corresponding label  of node $i$ \\
    $\bA$       & the adjacency matrix \\
    $\bar{\bA}$ & the augmented adjacency matrix \\ 
    $\widetilde{\bA}$   & learned feature-based matrix \\ 
    $\widehat{\bA}$     & $(1-\alpha)\bA+\alpha \widetilde{\bA}$ \\
    $\bar{\bD}$ & the diagonal matrix of $\bA+I$ \\
    $\widetilde{\bD}$ & the diagonal matrix of $\widetilde{\bA}$ \\
    $\Theta = (\bW_1, \bW_2)$   & parameters of two-layer GCN \\
    $\sigma$    & activation functions \\
    $\ba$       & trainable parameter with a sparsity constraint \\
    $\bM$  & trainable parameter used for projecting node embeddings into a latent space \\
    $\widehat{\mathcal{R}}(\mathcal{F})$    & the empirical Rademacher complexity\\
    $\mathfrak{R}_{m, n}$   & the transductive Rademacher complexity \\
    $\odot$     & the hadamard product \\
    $\|\ba\|_2$ & the standard norm in the Euclidean space \\
    $\|\bZ\|_2$, $\|\bZ\|_F$ & the spectral norm and Frobenius norm of $\bZ$ \\
       & \\
    \bottomrule
    \end{tabular}
\end{table}

\section{The Derivation of the Metric Function between Pairwise Representations} \label{sec:derivation}
Consider \cref{phixixj} , it can transform to the following
\begin{align} \label{transform}
\begin{split}
    \phi(\bx_i,\bx_j)&=\sqrt{(\ba\circ(\bx_i-\bx_j))^T\bM^T \bM \ba\circ(\bx_i-\bx_j)} 
                    =\sqrt{ (\bM\ba\circ(\bx_i-\bx_j))^T \bM\ba\circ(\bx_i-\bx_j) } \\
                    &=\sqrt{ (\bM\ba\circ\bx_i - \bM\ba\circ\bx_j)^T (\bM\ba\circ\bx_i - \bM\ba\circ\bx_j) } 
                    \\&
                    =\sqrt{ (\widetilde{\bx}_i - \widetilde{\bx}_j)^T (\widetilde{\bx}_i - \widetilde{\bx}_j)} 
                    = \| \widetilde{\bx}_i - \widetilde{\bx}_j \|_2,
\end{split}
\end{align}
where $\widetilde{\bx}=\bM\ba\circ\bx$ is a linear function. 
From the expressions above, we can consider \cref{phixixj} to be Euclidean distance. So, given parameter $W$ and $a$, we can transform feature $bX$  and then compute the Euclidean distance between nodes.

Given a metrix learning $\bD_{i j} = (\bx_i-\bx_j)^T(\bx_i-\bx_j)$, consider that
\begin{align*}
    \bD_{i j} &= (\bx_i-\bx_j)^T(\bx_i-\bx_j) = \bx_i^T \bx_i - 2\bx_i^T \bx_j + \bx_j^T \bx_j.
\end{align*}

Let $\bG_{i j}=\bx_i^T \bx_j$, so that $\bD_{i j}=\bG_{i i} - 2\bG_{i j} + \bG_{j j}$ and define $\mathbf H_{i j} = \bG_{i i}, \mathbf K_{i j}=\bG_{j j}$, $\mathbf H=\mathbf K^T$ in matrix 
\begin{align*}
    H &= 
    \begin{pmatrix}
          \bG_{1 1} & \bG_{1 1} & \cdots & \bG_{1 1} \\
          \bG_{2 2} & \bG_{2 2} & \cdots & \bG_{2 2} \\
          \vdots & \vdots & \vdots & \vdots \\
          \bG_{n n} & \bG_{n n} & \cdots & \bG_{n n} \\
    \end{pmatrix},
    \;
    K =
    \begin{pmatrix}
          \bG_{1 1} & \bG_{2 2} & \cdots & \bG_{n n} \\
          \bG_{1 1} & \bG_{2 2} & \cdots & \bG_{n n} \\
          \vdots & \vdots & \vdots & \vdots \\
          \bG_{1 1} & \bG_{2 2} & \cdots & \bG_{n n} \\
    \end{pmatrix},
    \;
    G =
    \begin{pmatrix}
          \bG_{1 1} & \bG_{1 2} & \cdots & \bG_{1 n} \\
          \bG_{2 1} & \bG_{2 2} & \cdots & \bG_{2 n} \\
          \vdots & \vdots & \vdots & \vdots \\
          \bG_{n 1} & \bG_{n 2} & \cdots & \bG_{n n} \\
    \end{pmatrix}.
\end{align*}

So a metric learning $\bD = \mathbf H + \mathbf K-2\bG$. Therefore, we can transform a for loop  to matrix multiplies using the function above
$$
    \widetilde{\bA}=\exp\Big(-\frac{\widetilde{\bD}}{2\sigma^2}\Big),
$$
where $\widetilde{\bD}$ is metric learning of $\phi^2(\bx_i, \bx_j)$. Given node feature $\bX$ and parameter $\bM, \ba$, first we calculate $\widetilde{\bX} = f(\bX)=\bM\ba\circ X$, 
then, let $\widetilde{\bG}=\widetilde{\bX}^T \widetilde{\bX} $, and $\widetilde{\mathbf H}_{i j} = \widetilde{\bG}_{i i}, \widetilde{\mathbf K}_{i j}=\widetilde{\bG}_{j j}$, and we can get $\widetilde{\bD} = \widetilde{\mathbf H} + \widetilde{\mathbf K} - 2\widetilde{\mathbf G}$. Finally, we can calculate $\widetilde{\bA} = \exp\big(-\nicefrac{\widetilde{\bD}}{2\sigma^2}\big)$.

\section{Theoretical Background and Technicalities} \label{sec:background}
In this section, we introduce some definitions and theoretical results necessary for our core idea's source.
We briefly recall the framework for transductive learning. Let $\mathcal{X} \triangleq\left\{\boldsymbol{x}_i\right\}_{i=1}^n$ be the domain with features $\boldsymbol{x}_i \in \mathbb{R}^d$ and $\{y_i\} \in \{\pm 1\}$ be the known label set. The goal of learning is to find a predictor $h$ which can minimise the generalisation error $\mathcal{L}_u(h) \triangleq \frac{1}{n-m} \sum_{i=m+1}^n \ell\left(h\left(x_i\right), y_i\right)$. In this way, the generalization error bound for graph-based transduction takes the form as
$$
    \mathcal{L}_u(h) \leq \widehat{\mathcal{L}}_m(h)+\text { complexity term},
$$
where $\widehat{\mathcal{L}}_m(h) \triangleq \frac{1}{m} \sum_{i=1}^m \ell\left(h\left(x_i\right), y_i\right)$ is the empirical error of $h$ and 
$\mathcal{L}_n(h) \triangleq \frac{1}{n} \sum_{i=1}^n \ell\left(h\left(x_i\right), y_i\right)$. The complexity term is typically characterized using learning-theoretic terms such as VC dimension and Rademacher complexity\citep{el2009transductive, tolstikhin2016minimax}. 

The following definition of the Transductive Rademacher complexity avoids the triviality of VC Dimension based error bounds and extends inductive Rademacher complexity by considering the unobserved instances.

\begin{definition} [Transductive Rademacher complexity, \citealt{el2009transductive}]
Let $\mathcal{V} \subseteq \mathbb{R}^n, p \in$ $[0,1/2]$ and $m$ labeled points. Let $\boldsymbol{\sigma}=\left(\sigma_1, \ldots, \sigma_n\right)^T$ be a vector of i.i.d. random variables, where $\sigma_i$ takes value $+1$ or $-1$, $P(\sigma_i=1)=P(\sigma_i=-1)=p$, and $P(\sigma_i=0)=1-2p$. The transductive Rademacher complexity of $\mathcal{V}$ is 
$$
    \mathfrak{R}_{m, n}(\mathcal{V}) \triangleq\left(\frac{1}{m}+\frac{1}{n-m}\right) \cdot \underset{\sigma}{\mathbb{E}}\left[\sup _{\mathbf{v} \in \mathcal{V}} \boldsymbol{\sigma}^{\top} \mathbf{v}\right] .
$$
\end{definition}

Next, the following result derives a bound for the TRC of $K$-layer GNNs and states the corresponding generalization error bound. 

\begin{thm}[Generalization error bound for GNNs using TRC, \citealt{esser2021learning}] \label{thm:2}
Assume $\mathcal{H}_{\mathcal{G}}^{\phi, \beta, \omega} \subseteq \mathcal{H}_{\mathcal{G}}^\phi$ where the trainable parameters satisfy $\left\|\boldsymbol{b}_k\right\|_1 \leq \beta$ and $\left\|\boldsymbol{W}_k\right\|_{\infty} \leq \omega$ for every $k \in[K]$. The transductive Rademacher complexity (TRC) of the restricted hypothesis class is bounded as
\begin{align*}
    \mathfrak{R}_{m, n}\left(\mathcal{H}_{\mathcal{G}}^{\phi, \beta, \omega}\right) 
    \leq& \frac{c_1 n^2}{m(n-m)}\left({\sum}_{k=0}^{K-1} c_2^k\|\boldsymbol{S}\|_{\infty}^k\right)
    +c_3 c_2^K\|\boldsymbol{S}\|_{\infty}^K\|\boldsymbol{S} \boldsymbol{X}\|_{2 \rightarrow \infty} \sqrt{\log n},
\end{align*}
where $c_1 \triangleq 2 L_\phi \beta, c_2 \triangleq 2 L_\phi \omega, c_3 \triangleq L_\phi \omega \sqrt{2 / d}$ and $L_\phi$ is Lipschitz constant for activation $\phi$.

Following \citep{el2009transductive}, the bound leads to a generalization error bound. For any $\delta \in(0,1)$ and $h \in \mathcal{H}_{\mathcal{G}}^{\phi, \beta, \omega}$, the generalisation error satisfies
\begin{align*}
    \mathcal{L}_u(h)-\widehat{\mathcal{L}}_m(h) 
    \leq& \mathfrak{R}_{m, n}\left(\mathcal{H}_{\mathcal{G}}^{\phi, \beta, \omega}\right)+c_4 \frac{n \sqrt{\min \{m, n-m\}}}{m(n-m)}
    +c_5 \sqrt{\frac{n}{m(n-m)} \ln \left(\frac{1}{\delta}\right)},
\end{align*}
with probability $1-\delta$, where $c_4, c_5$ are absolute constants such that $c_4<5.05$ and $c_5<0.8$.
\end{thm}
Note that the bound in \cref{thm:2} depends on the alignment between the feature and graph information.

\section{Proof of \cref{lower}: Lower Bound of Rademacher Complexity}
\label{sec:proof}
\begin{proof}[Proof of \cref{lower}]
By definition of the Rademacher complexity, it is enough to lower bound the complexity of some subset of $\mathcal{F}_{D,R}$, denoted by $\mathcal{F}'$. In particular, we focus on the class $\mathcal{F}_{D,R}'$ of graph neural networks over $\mathbb{R}$ of the form
\begin{align*}
    \mathcal{F}_{D,R}':=\Big\{f(\bx_i)
    &=\sigma\Big({\sum}_{t=1}^kw_t^{(2)}{\sum}_{v=1}^n\bar{\bA}_{iv}
    \times \sigma\big({\sum}_{j\in N(v)}\bar{\bA}_{vj} \big\langle\bx_j, \bw_t^{(1)} \big\rangle \big) \Big),\;i\in[m], 
    \\ & 
    \bW^{(1)}=(\bw^{(1)},{\bf 0},...,{\bf 0}),\; 
    \bw^{(2)}=(w^{(2)},{\bf 0});\; 
    \|\bw^{(1)}\|_{2}\leq R,\,|w^{(2)}|\leq D \Big\},
\end{align*}

where we choose  $\bW^{(1)}=(\bw^{(1)},{\bf 0},...,{\bf 0})$ so that only  the first column vector could be nonzero. In this case, $\|\bW^{(1)}\|_F=\|\bw^{(1)}\|_{2}\leq R$ holds. Similarly, we only allow $\bw^{(2)}$ to vary in the first coordinate to simplify the proof. Furthermore, we take the linear activation $\sigma(s)=ls$ as our choice.
In this setup, it holds that
\begin{align} \label{lowerb}
    \widehat{\mathcal{R}}(\mathcal{F}_{D,R}')
    &\geq l^2\mathbb{E}_{\bepsilon}\sup_{\|\bw^{(1)}\|_2\leq R,\,|w^{(2)}|\leq D}\frac{1}{m}\Big| {\sum}_{i=1}^m\epsilon_i\Big( w^{(2)}{\sum}_{v=1}^n\bar{\bA}_{iv}
    \times \big({\sum}_{j\in N(v)}\bar{\bA}_{vj} \langle \bx_{j},\bw^{(1)}\rangle\big)\Big)\Big| \nonumber\\
    &=\frac{l^2D}{m} \mathbb{E}_{\bepsilon}\sup_{\|\bw^{(1)}\|_2\leq R}\Big|\Big\langle {\sum}_{i=1}^m\epsilon_i\Big( {\sum}_{v=1}^n\bar{\bA}_{iv}\times \big({\sum}_{j\in N(v)}\bar{\bA}_{vj} \bx_{j} \big)\Big),\,\bw^{(1)}\Big\rangle\Big| \nonumber\\
    &=\frac{l^2RD}{m} \mathbb{E}_{\bepsilon}\Big\| {\sum}_{i=1}^m\epsilon_i\Big( {\sum}_{v\in N(i)}\bar{\bA}_{iv}\times \big({\sum}_{j\in N(v)}\bar{\bA}_{vj} \bx_{j} \big)\Big)\Big\|_2, 
\end{align}
where the last step follows from the neighbor representation of graph shift operators, as well as the equivalent form of the $L_2$-norm, that is, $\|{\bf s}\|_2=\sup_{\|\bw\|_2=1}\langle  {\bf s},\bw\rangle$. Let ${\bf e_1}=(1,0,...,0)$ denote the standard unit vector in $\mathbb{R}^d$, and we assume that all the input data have the specific form $\bx_{j}=B{\bf e_1}$ for all $j\in [n]$. 
Then
\begin{align}\label{remx}
\begin{split}
    &\Big\| {\sum}_{i=1}^m\epsilon_i\Big( {\sum}_{v\in N(i)}\bar{\bA}_{iv}\big({\sum}_{j\in N(v)}\bar{\bA}_{vj} \bx_{j} \big)\Big)\Big\|_2
    =B\Big| {\sum}_{i=1}^m\epsilon_i\Big( {\sum}_{v\in N(i)}\bar{\bA}_{iv} \big({\sum}_{j\in N(v)}\bar{\bA}_{vj} \big)\Big)\Big|.
\end{split}
\end{align}
Note that, the exchange of summation leads to the following equality,
$$
{\sum}_{i=1}^m\epsilon_i\Big( \sum_{v\in N(i)}\bar{\bA}_{iv}\times \big({\sum}_{j\in N(v)}\bar{\bA}_{vj} \big)\Big)={\sum}_{k=1}^q{\sum}_{t=1}^q\bar{\bA}_{kt}\Big({\sum}_{i=1}^m\epsilon_i\bar{\bA}_{ik}\Big).  
$$
Suppose that the term $\sum_{t=1}^q\bar{\bA}_{kt}$ is invariant with $k$, denoted by $h_q(\bL)$.  Then
$$
{\sum}_{i=1}^m\epsilon_i\Big({\sum}_{v\in N(i)}\bar{\bA}_{iv}\times \big({\sum}_{j\in N(v)}\bar{\bA}_{vj} 
\big)\Big)=h_q(\bL){\sum}_{i=1}^m\epsilon_i\Big({\sum}_{k=1}^q\bar{\bA}_{ik}\Big).  
$$
Hence, this together with \cref{remx} yields that 
\begin{align}\label{eachangeq}
\begin{split}
    &\mathbb{E}_{\bepsilon}\Big\|
    {\sum}_{i=1}^m\epsilon_i\Big( {\sum}_{v\in N(i)}\bar{\bA}_{iv} \big({\sum}_{j\in N(v)}\bar{\bA}_{vj} \bx_{j}\big)\Big)\Big\|_2
    =h^2_q(\bL)\mathbb{E}_{\bepsilon}\Big|{\sum}_{i=1}^m\epsilon_i\Big|=h^2_q\sqrt{m}.
\end{split}
\end{align}
Moreover, by  our choice of $\bx_{j}$ for all $j$ as above, we can check that
$$
|h_q(\bL)|=|\langle \bar{\bA}_{\cdot k}, {\bf 1}\rangle|=\big\|\widetilde{\bX}_q\bar{\bA}_{\cdot k} \big\|_2.
$$
As a consequence, combining \cref{lowerb,remx,eachangeq}, we obtain
\begin{align*}
\widehat{\mathcal{R}}(\mathcal{F}_{D,R}')\geq
\frac{l^2BRD}{\sqrt{m}}\min_{k\in [q]}\Big\{ \big\|\widetilde{\bX}_q\bar{\bA}_{\cdot k} \big\|_2{\sum}_{t=1}^q\bar{\bA}_{kt}\Big\}.
\end{align*}
This completes the proof of \cref{lower}.
\end{proof}

\end{appendices}

\end{document}